\pdfoutput=1

\documentclass[11pt]{article}

\usepackage{emnlp2022}
\usepackage{times}
\usepackage{latexsym}
\usepackage{url}
\usepackage{arydshln}
\usepackage{graphicx}
\usepackage{amsmath,amsthm,amssymb}
\usepackage{array}
\usepackage{booktabs}
\usepackage{makecell}
\usepackage{multirow}
\usepackage{cleveref}
\usepackage{enumitem}
\usepackage{color}
\crefname{section}{§}{§§}
\Crefname{section}{§}{§§}
\usepackage[T1]{fontenc}

\usepackage[utf8]{inputenc}

\usepackage{microtype}

\title{CARE: Causality Reasoning for Empathetic Responses by\\Conditional Graph Generation}

\author{
Jiashuo WANG, Yi CHENG, Wenjie LI\thanks{\ \ Corresponding author.}\\
Hong Kong Polytechnic University\\
\tt \{csjwang,csycheng,cswjli\}@comp.polyu.edu.hk\\ 
}

\begin{document}
\maketitle
\begin{abstract}
Recent approaches to empathetic response generation incorporate emotion causalities to enhance comprehension of both the user's feelings and experiences. However, these approaches suffer from two critical issues. 
First, they only consider causalities between the user's emotion and the user's experiences, and ignore those between the user's experiences. 
Second, they neglect interdependence among causalities and reason them independently.
To solve the above problems, we expect to reason all plausible causalities interdependently and simultaneously, given the user's emotion, dialogue history, and future dialogue content. Then, we infuse these causalities into response generation for empathetic responses. 
Specifically, we design a new model, i.e., the Conditional Variational Graph Auto-Encoder (CVGAE), for the causality reasoning, and adopt a multi-source attention mechanism in the decoder for the causality infusion. We name the whole framework as \textbf{CARE}\footnote{The implementation of CARE is publicly available at \url{https://github.com/wangjs9/CARE-master}.}, abbreviated for \textbf{CA}usality \textbf{R}easoning for \textbf{E}mpathetic conversation.
Experimental results indicate that our method achieves state-of-the-art performance. 
\end{abstract}

\section{Introduction}
Empathy is the capability to perceive, understand and respond to another individual's feelings, experiences and situation \cite{paiva2017empathy,decety2004functional}. It is composed of two aspects \cite{davis1983measuring}, which are (\romannumeral1) affection, i.e., emotion understanding and appropriate emotional reaction \cite{hoffman2001empathy}, and (\romannumeral2) cognition, i.e., comprehension and reasoning of the other's experiences and situation \cite{preston2002empathy}. 

\begin{figure}
    \centering
    \includegraphics[width=\linewidth]{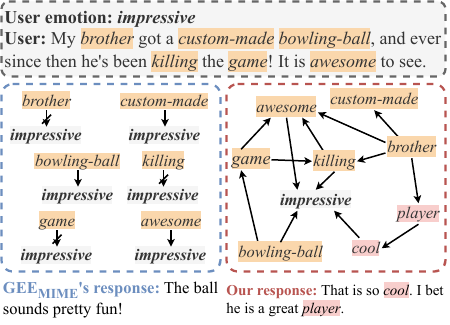}
    \caption{Causality reasoning results of GEE$_{\text{MIME}}$ \cite{Kim:2021:empathy} and our proposed method in a real case. Arrows indicate relations from cause to effect, while strikeout arrows indicate no causal relations. GEE$_{\text{MIME}}$ detects only direct causes and effects of the user's emotion independently, while ours extends the causality scope and reasons causalities interdependently.}
    \label{fig:CVGAE}
\end{figure}

Earlier work on empathetic response generation merely pays attention to affection \cite{moel,majumder2020mime,li2019empgan}. Consequently, their models lack understanding of the user's experiences, resulting in very weak empathy.
Most recent studies begin to consider both affection and cognition by incorporating emotion cause and effect \cite{wang2021empathetic,gao-etal-2021-improving-empathetic,Kim:2021:empathy,CEM2021}. Despite notable improvement, their methods suffer from two critical problems. 
First, they only consider causalities between the user's emotion and the user's experiences, which are just part of cognition. Causalities between experiences also contribute to the comprehension of experiences. For the case in \Cref{fig:CVGAE}, although \textit{brother} does not cause \textit{impressive} directly, it is the subject causing what impresses the user. Therefore, \textit{brother} should be considered in the causal information for response generation.
Second, these methods reason causalities independently and ignore interdependence among these causalities, leading to low-fidelity causality detection. 
As shown in \Cref{fig:CVGAE}, GEE$_{\text{MIME}}$, one of these methods, fails to reason \textit{killing} $\rightarrow$ \textit{impressive}, since \textit{killing} itself ordinarily is the cause or effect of a negative emotion. However, this causality is reasonable when simultaneously considering other causalities including \textit{game} $\rightarrow$ \textit{killing}, as our proposed method models.
Due to the above two problems, these previous methods always misunderstand feelings and experiences of the user, impeding empathetic expression in responses.

To solve these problems, we propose to reason all plausible causalities, i.e., causalities stated explicitly in the dialogue history and probably in the future dialogue, interdependently and simultaneously by formulating the reasoning as a conditional graph generation task.
Specifically, we aim to generate a causal graph\footnote{Each node is a word to represent the user's feelings and experiences, and each edge indicates a causal relationship between two nodes.} containing all plausible causalities conditioned on the user's emotion, dialogue history, and predicted future dialogue content.
Inspired by the Variational Graph Auto-Encoder (VGAE) \cite{kipf2016variational}, we design a Conditional Variational Graph Auto-Encoder (CVGAE), which uses latent variables for conditional structure prediction, to accomplish causality reasoning.
Accordingly, the model is expected to have a deeper understanding of the user's feelings and experiences.
In addition, some feelings and experiences, which are not explicitly stated in dialogue history but contribute to response generation, can be inferred in this process as shown in \Cref{fig:CVGAE}.

In this paper, we propose a novel empathetic response generation model, called \textbf{CARE} (\textbf{CA}usality \textbf{R}easoning for \textbf{E}mpathetic conversation).
CARE reasons all plausible causalities by CVGAE, and infuses them into response generation by a multi-source attention mechanism in the decoder. 
In addition, we adopt multi-task learning to integrate causality reasoning and response generation during training.
The experimental results on the \textsc{EmpatheticDialogues} \cite{empatheticdialogues} benchmark suggest that our method improves the model's understanding of user's feelings and experiences, and \textbf{CARE} achieves state-of-the-art performance on empathetic response generation.

Our main contributions are three-fold:
\begin{enumerate}[leftmargin=*,label=\arabic*).]
    \item We propose to reason all plausible causalities in empathetic conversation interdependently and simultaneously for a deep understanding of the user's feelings and experiences.
    \item We turn causality reasoning into a conditional graph generation task, and introduce CVGAE, which uses latent variables for conditional structure prediction, to achieve the reasoning.
    \item We design CARE, which augments empathetic response generation with causality reasoning, and prove its outstanding performance on the \textsc{EmpatheticDialogues} benchmark.
\end{enumerate}
\section{Related Work}
Since empathy is a critical character for social chatting systems \cite{sharma2020computational,perez2017understanding}, many studies have contributed to empathetic response generation. Earlier work mainly focuses on the affective aspect of empathy. 
MoEL \cite{moel} adopts a mixture of experts architecture to combine outputs from different decoders, each of which represents one emotion. 
Based on the idea of emotion mixture, MIME \cite{majumder2020mime} takes emotion polarity (positive or negative) into account. Moreover, it uses emotion stochastic sampling and emotion mimicry to generate empathetic responses. 
\citet{li2019empgan} propose to capture nuances of emotion at the token-level for decoding. Moreover, an adversarial learning framework is leveraged to involve user feedback.

Having realized that ignorance of cognition impedes empathy in conversation, some recent methods involve both affection and cognition by incorporating emotion causes and effects. 
\citet{wang2021empathetic} incorporate emotion causes into empathetic response generation by multi-hop reasoning from emotion causes to emotion states. 
\citet{gao-etal-2021-improving-empathetic} identify emotion causes from dialogue context, and use gates at the decoder to control the involvement of these emotion causes in the response generation.
\citet{Kim:2021:empathy} emphasize emotion causes in dialogue context by a rational speech act framework. 
These three methods identify emotion causes via a classifier, which detects whether there is a causal relationship between a conversation fragment and an emotion statement or word each time.
CEM \cite{CEM2021} uses COMET, an if-then commonsense generator, to generate causes and effects of user experiences, and refines dialogue context with them for response generation. However, all these methods obtain causalities independently.

\section{Preliminary}
\subsection{Transformer-based Response Generation}
The response generation model is built upon the vanilla transformers \cite{transformers}, which generates the response $R$ given dialogue context $C$ as input in an encoder-decoder manner. 
The encoder encodes the dialogue context and generates the context hidden state. That is:
\begin{equation}\label{eqn:encoder}
    E_{out} = \text{TRS}_{\text{enc}}(C),
\end{equation}
$E_{out} \in \mathbb{R}^{|C|\times d}$, where $d$ is the hidden size. 
The decoder takes the right shifted response as input and generates the response. Typically, the whole decoder includes $L_{dec}$ decoder layers, each consisting of three sub-layers.
The first one, i.e., the self-attention sub-layer, computes a representation of the input sequence:
\begin{equation}\label{eqn:self}
\begin{gathered}
    \hat{H}=\text{MultiHead}(H_{in}, H_{in}, H_{in}), \\
    H^{(self)}_{out}=\text{LayerNorm}(\hat{H}+H_{in}),
\end{gathered}
\end{equation}
where $H_{in}$ is the embedding right shifted response for the first decoder layer, and is output of the $(l-1)\text{-th}$ decoder layer for the $l\text{-th}$ decoder layer.
Then the decoder attends to the dialogue context by a cross-attention sub-layer:
\begin{equation}\label{eqn:cross}
\begin{gathered}
    H_{in}=H^{(self)}_{out},\\
    \hat{H}=\text{MultiHead}(H_{in}, E_{out}, E_{out}), \\
    H^{(cross)}_{out}=\text{LayerNorm}(\hat{H}+H_{in}). 
\end{gathered}
\end{equation}
The output of the $l\text{-th}$ decoder layer is obtained by the feed-forward sub-layer: 
\begin{equation}\label{eqn:ffn}
\begin{gathered}
    H_{in}=H^{(cross)}_{out},\\
    H^{(ffn)}_{out}=\text{LayerNorm}(\text{FFN}(H_{in})+H_{in}).
\end{gathered}
\end{equation}
Finally, we apply linear transformation and a softmax operation on the output of the $L_{dec}$ decoder layer to predict token probability distribution at each token position $t$:
\begin{equation}\label{eqn:prob}
    P_t=softmax(H^L_{out,t}W_o+b_o),
\end{equation}
where $H^L_{out,t}$ is the final output for the $t$-th token; $W_o \in \mathbb{R}^{d\times d_{vocab}}$ and $b_o \in \mathbb{R}^{d_{vocab}}$ are parameters, and $d_{vocab}$ is the vocabulary size. 

\subsection{Variational Graph Auto-Encoder}\label{sec:VGAE}
Our proposed causality reasoning module, i.e., CVGAE, is based on VGAE \cite{kipf2016variational}. Given an undirected graph $\mathcal{G}=(\mathcal{V}, \mathcal{E})$ with its adjacency matrix $\mathbf{A}$, VGAE generates graph latent variables by an inference model, and reconstructs the adjacency matrix by a generative model.

\paragraph{Inference Model} 
The inference model encodes $\mathcal{G}$, and generates graph latent variables $\mathbf{Z}=\{\mathbf{z}_1, \dots, \mathbf{z}_{|\mathcal{V}|}\}$ by a recognition net $q(\mathbf{Z}|\mathcal{V},\mathbf{A})$. Each graph latent variable $z_i$ is obtained by:
\begin{equation}\label{eqn:g_latent}
\begin{aligned}
    & q(\mathbf{z}_i|\mathcal{V},\mathbf{A})=\mathcal{N}(\mathbf{z}_i|\mu_i,{\mathbf{\sigma}^2_i}),\\
    \text{with} \quad & \mu=\text{GCNLayer}_{\mu}(\mathcal{H}_{\mathcal{V}}, \mathbf{A}), \\
    \text{and} \quad & \text{log}\sigma=\text{GCNLayer}_{\sigma}(\mathcal{H}_{\mathcal{V}}, \mathbf{A}).
\end{aligned}
\end{equation}
Here, $\mathcal{N}$ is a sampling function, which follows the Gaussian distribution. $\mu$ is the matrix of the mean vectors $\mu_i$; $\text{log}\sigma$ is the matrix of log-variance vectors $\text{log}\sigma_i$. In particular, $\mathcal{H}_{\mathcal{V}}$ is a shared hidden state obtained by:
\begin{equation}\label{eqn:shared_hidden}
    \mathcal{H}_{\mathcal{V}}=\text{GCNLayer}_{\text{h}}(\mathcal{V}, \mathbf{A}).
\end{equation}

\paragraph{Generative Model} 
The generative model reconstructs the adjacency matrix by an inner product between latent variables:
\begin{equation}\label{eqn:generative_A}
\begin{aligned}
    & p(\hat{\mathbf{A}}|\mathbf{Z})=\prod_{i=1}^{|\mathcal{V}|}\prod_{j=1}^{|\mathcal{V}|}p(\hat{\mathbf{A}}_{ij}|\mathbf{z}_i,\mathbf{z}_j),\\
    \text{with} \quad & p(\hat{\mathbf{A}}_{ij}=1|\mathbf{z}_i\mathbf{z}_j)=\text{sigmoid}(\mathbf{z}_i^\top\mathbf{z}_j).
\end{aligned}
\end{equation}

\paragraph{Inference Stage}
At the inference stage, adjacency matrix $\mathbf{A}$ is unavailable. Therefore, we replace $q(\mathbf{Z}|\mathcal{V},\mathbf{A})$ with a prior net $p(\mathbf{Z})$, which is parameterized by a Gaussian distribution: $p(\mathbf{z}_i)=\mathcal{N}(\mathbf{z}_i|0,1)$, to infer $\mathbf{Z}$. Then, we use the same generative model to generate the adjacency matrix.

\paragraph{Objective}
VGAE is optimized by maximizing:
\begin{equation}\label{eqn:VGAE_loss}
    \begin{aligned}
        \mathcal{L}=&\mathbb{E}_{q(\mathbf{Z}|\mathcal{V}, \mathbf{A})}[\text{log}p(\hat{\mathbf{A}}|\mathbf{Z})] \\ &\qquad - \text{KL}[q(\mathbf{Z}|\mathcal{V}, \mathbf{A})||p(\mathbf{Z})],
    \end{aligned}
\end{equation}
where $KL[q(\cdot)|p(\cdot)]$ is the Kullback-Leibler divergence between $q(\cdot)$ and $p(\cdot)$.

\section{Method}
\begin{figure*}[t]
    \centering
    \includegraphics[width=\textwidth]{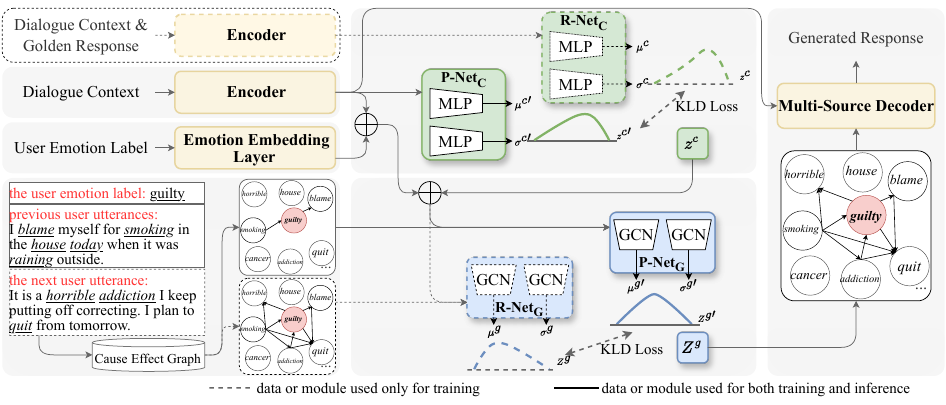}
    \caption{The overview of our proposed framework. The solid lines represent modules or data used for both posterior and prior computation, while the dot lines represent modules or data used only for posterior computation.}
    \label{fig:overview}
\end{figure*}
\Cref{fig:overview} presents an overview structure of our proposed model CARE. It first reasons all plausible causalities interdependently by generating a causal graph. Specifically, we use CVGAE to generate this graph under the condition of the user's emotion, dialogue history, and predicted future dialogue content. Notably, CVGAE works differently at the training and inference stages: it reconstructs a posterior causal graph (by \textit{R-Net}$_G$) with this posterior causal graph as input during training, while generates a posterior causal graph (by \textit{P-Net}$_G$) with a prior causal graph as input during inference.
The prior causal graph contains causalities explicitly mentioned in previous user utterances, while the posterior one contains additional causalities in the next user utterance. 
Then, CARE infuses causalities in the reasoned causal graph into response generation by multi-source attention at the decoder.

\subsection{Graph Construction}
As mentioned, we need a prior causal graph $\mathcal{G}_{prior}=(\mathcal{V}, \mathcal{E}_{prior})$ and a posterior causal graph $\mathcal{G}_{post}=(\mathcal{V}, \mathcal{E}_{post})$, for causality reasoning at inference and training stage, respectively.
We construct them with the assistance of a causal knowledge graph, i.e., Cause Effect Graph (CEG) \cite{ijcai2020-guided}.
These two graphs share the same node set, which theoretically contains all nodes in CEG. 
However, for effectiveness, we only consider those among a certain set of nodes $\mathcal{V}$, which contains emotion label word, words appearing in previous user utterances, and one-hop neighbors of above two kinds of words. 
The edge sets of these two graphs are different.
$\mathcal{E}_{prior}$ contains causal relationships in previous user utterances, while $\mathcal{E}_{post}$ also contains those in the next utterances. In specific, we collect $\mathcal{E}_{prior}$ and $\mathcal{E}_{post}$ according to following rules.
For any couple nodes in $\mathcal{V}$ having a relationship in CEG, if both nodes are covered by the user emotion label word and words in previous user utterances, we add the relationship into $\mathcal{E}_{prior}$; if both nodes are covered by the user emotion label word and words in previous and next user utterances, we add the relationship into $\mathcal{E}_{post}$.

\subsection{Conditional Variational Graph Auto-Encoder (CVGAE)}
We design a novel structure CVGAE to generate a (posterior) causal graph for causality reasoning.
As an extension of VGAE, CVGAE works in a similar manner (\Cref{sec:VGAE}). In particular, it generates graphs latent variables for graph reconstruction under some conditions, including a context condition, an emotion condition, and a context latent variable.

\paragraph{Context and Emotion Conditions}
The context condition is expected to provide information of dialogue context $C$, thus it is derived from the encoder output. Following \cite{Wang2019TCVAETC}, we use multi-head attention to perform it. That is:
\begin{equation}
    \mathbf{c}_{\text{ctx}} = \text{MultiHead}(v_{\text{rand}}, E_{out}, E_{out}),\label{eqn:con_repr}
\end{equation}
where $E_{out} \in \mathbb{R}^{|C|\times d}$ is the encoder output computed by $\text{TRS}_{\text{enc}}(C)$ in \Cref{eqn:encoder}; $v_{\text{rand}} \in \mathbb{R}^{1\times d}$ is a randomly initialized vector and is regarded as a single query for multi-head attention. 

The emotion condition is expected to provide information of the user emotion $e$. Accordingly, we define the emotion embedding $E^{emo}\in \mathbb{R}^d$ which converts an emotion label into embeddings. The emotion condition is formulated as:
\begin{equation}
    \mathbf{c}_{\text{emo}} = E^{emo}(e).
\end{equation}

\paragraph{Context Latent Variable}
We use a context latent variable $z_c$ to provide information from the future dialogue. This variable is generated by a contextual recognition net (\textit{R-Net}$_C$ in \Cref{fig:overview}) with dialogue context $C$ and the golden response $R$ as input:
\begin{equation}
    q_c(\mathbf{z}^{\text{c}}|C,R)=\mathcal{N}(\mathbf{z}^{\text{c}}|\mu^c,{\sigma^c}^2).
\end{equation}
Here $\mu^c=\text{MLP}_\mu(\mathbf{c}_{\text{lant}})$ is the mean vector and $\text{log}\sigma_c=\text{MLP}_\sigma(\mathbf{c}_{\text{lant}})$ is the log-variance vector, where $\mathbf{c}_{\text{lant}}$ is accessed similar to \Cref{eqn:con_repr}:
\begin{gather}
    E_{rep} = \text{TRS}_{\text{enc}}(C\oplus R),\\
    \mathbf{c}_{\text{lant}} = \text{MultiHead}(v_{\text{rand}}, E_{rep}, E_{rep}).
\end{gather}

\paragraph{Graph Latent Variables} We generate graph latent variables $\mathbf{Z}^g$ by a recognition net (\textit{R-Net}$_G$ in \Cref{fig:overview}): $q_g(\mathbf{Z}^g|\mathcal{V},\mathbf{A}_{post},c_{\text{cond}})$, where $\mathbf{A}_{post}$ is the adjacency matrix of $\mathcal{G}_{post}$. This process is similar to that of VGAE, i.e., \Cref{eqn:g_latent,eqn:shared_hidden}.
\begin{equation}\label{eqn:latent_variable_G}
\begin{aligned}
    q(\mathbf{z}^g_i|\mathcal{V},\mathbf{A}_{post},c_{\text{cond}})=\mathcal{N}(\mathbf{z}^g_i|\mu^g_i,{\mathbf{{\sigma}^g}^2_i}),\\
    \text{with} \quad \mu^g=\text{GCNLayer}_{\mu}(\mathcal{H}_{\mathcal{V}}, \mathbf{A}_{post}), \\
    \text{and} \quad \text{log}\sigma^g=\text{GCNLayer}_{\sigma}(\mathcal{H}_{\mathcal{V}}, \mathbf{A}_{post}).
\end{aligned}
\end{equation}
The shared hidden state $\mathcal{H}_{\mathcal{V}}$ is generated with attention to the concatenation of $\mathbf{c}_{\text{ctx}}$, $\mathbf{c}_{\text{emo}}$, and $\mathbf{z}^{\text{c}}$:
\begin{equation}\label{eqn:shared_hidden_G}
\begin{aligned}
    c_{\text{cond}}=\mathbf{c}_{\text{ctx}}\oplus \mathbf{c}_{\text{emo}}\oplus
    \mathbf{z}^{\text{c}},\\
    \hat{\mathcal{H}}_{\mathcal{V}}=\text{GCNLayer}_{\text{h}}(\mathcal{V}, \mathbf{A}_{post}),\\
    \mathcal{H}_{\mathcal{V}} = \text{MultiHead}(\hat{\mathcal{H}}_{\mathcal{V}},c_{\text{cond}},c_{\text{cond}}).
\end{aligned}
\end{equation}

\paragraph{Causal Relation Generation}
With graph latent variables $\mathbf{Z}^g$, we reconstruct the posterior causal graph, i.e., the matrix adjacency $\hat{\mathbf{A}}$ by \Cref{eqn:generative_A}. Then we select top-$k$ relationships from the reconstructed graph according to their probability, denoted as $\mathcal{R}=(r_1,\dots,r_k)$, where $r_i$ is the sum of the head and tail node embeddings.

\paragraph{Inference Stage}
During inference, $R$ (the golden response) and $\mathbf{A}_{post}$ are unavailable, thus we use a prior net $p_g({\mathbf{Z}^g}^{\prime}|\mathcal{V},\mathbf{A}_{prior},c_{\text{cond}}^{\prime})$  (\textit{P-Net}$_G$ in \Cref{fig:overview}) to approach $q_g(\mathbf{Z}^g)$, i.e., \Cref{eqn:latent_variable_G,eqn:shared_hidden_G}. $\mathbf{A}_{prior}$ is $\mathcal{G}_{prior}$'s adjacency matrix, and $c_{\text{cond}}^{\prime}=\mathbf{c}_{\text{ctx}}\oplus \mathbf{c}_{\text{emo}}\oplus {\mathbf{z}^{\text{c}}}^{\prime}$, where ${\mathbf{z}^{\text{c}}}^{\prime}$ is obtained by a contextual prior net (\textit{P-Net}$_C$ in \Cref{fig:overview}):
\begin{equation}
    p_c({\mathbf{z}^{\text{c}}}^{\prime}|C)=\mathcal{N}({\mathbf{z}^{\text{c}}}^{\prime}|{\mu^c}^{\prime},{{\sigma^c}^{\prime}}^2),
\end{equation}
with ${\mu^c}^{\prime}=\text{MLP}_{\mu^{\prime}}(\mathbf{c}_{\text{ctx}})$, $\text{log}{\sigma^c}^{\prime}=\text{MLP}_{\sigma^{\prime}}(\mathbf{c}_{\text{ctx}})$.

\subsection{Graph-Infused Response Generation}
To infuse the reasoned $\mathcal{R}$ into generation, we enable the decoder to attend to both dialogue context and the causal graph (\textit{Multi-Source Decoder} in \Cref{fig:overview}). In particular, we slightly modify the cross-attention sub-layer of the original decoder, i.e., \Cref{eqn:cross}, with our multi-source attention mechanism.
Therefore, the output after this modified sub-layer is computed by:
\begin{equation}
\begin{gathered}
    \hat{H}^C=\text{MultiHead}(H^{(cross)}_{in}, E_{out}, E_{out}), \\
    \hat{H}^{\mathcal{R}}=\text{MultiHead}(H^{(cross)}_{in}, \mathcal{R}, \mathcal{R}),\\
    \hat{H}=(\hat{H}^C\oplus\hat{H}^{\mathcal{R}})W_{multi},\\
    H^{(cross)}_{out}=\text{LayerNorm}(\hat{H}+H^{(cross)}_{in}),
\end{gathered}
\end{equation}
where $E_{out} \in \mathbb{R}^{|C|\times d}$ is the encoder output, $W_{multi}\in \mathbb{R}^{2d\times d}$ is a group of linear transformation parameters, and $H_{in}$ is the output of the self-attention sub-layer of the decoder computed by \Cref{eqn:self}. Notably, the reset of the original decoder, i.e, \Cref{eqn:self,eqn:ffn,eqn:prob}, remains the same.
In this way, we generate the final response.

\subsection{Training Objective}
We optimize the model with multi-task learning to further integrate the causality reasoning and the graph-infused response generation.
For the causality reasoning, we consider graph reconstruction accuracy and similarity between posterior and prior distribution. Similar to \Cref{eqn:VGAE_loss}, the corresponding loss can be calculated by:
\begin{equation}
    \begin{aligned}
        \mathcal{L}_{r}=&\mathbb{E}_{q_g(\mathbf{Z}^g|\mathcal{V}, \mathbf{A}_{post},c_{\text{cond}})}[\text{log}p(\hat{\mathbf{A}}|\mathbf{Z}^g)] \\ &-\text{KL}[q_g(\mathbf{Z}^g)||p_g({\mathbf{Z}^g}^{\prime})]\\ &-\text{KL}[q_c(\mathbf{z}^{\text{c}})||p_c({\mathbf{z}^{\text{c}}}^{\prime})].
    \end{aligned}
\end{equation}
The response generation loss is calculated by:
\begin{equation}
    \mathcal{L}_{g}=\prod_{t=1}^{|R|}P_t,
\end{equation}
where $P_t$ is obtained by \Cref{eqn:prob}.
Finally, we train CARE by maximizing $(\mathcal{L}_{r}+\mathcal{L}_{g})$.

\section{Experiments}
\subsection{Dataset}
We conduct our experiments on \textsc{EmpatheticDialogues}\footnote{\url{https://github.com/facebookresearch/EmpatheticDialogues}} \cite{empatheticdialogues}. It contains 25k crowdsourced one-on-one conversations, each of which is developed based on a particular emotion. There are $32$ emotion categories distributed in a balanced way. Following its original division, we adopt approximately $80\%$, $10\%$, and $10\%$ of the dataset for training, validation, and testing.

\subsection{Comparison Models}
We select seven models for comparison according to some special considerations. Three models that merely consider the affective aspect of the empathy are selected. They are:
\paragraph{MoEL}\footnote{\url{https://github.com/HLTCHKUST/MoEL}} \cite{moel}: This model leverages a mixture of expert architecture to combine outputs from several decoders, each of which pays attention to a unique emotion type.
\paragraph{MIME}\footnote{\url{https://github.com/declare-lab/MIME}} \cite{majumder2020mime}: Based on MoEL's idea of emotion mixture, this model takes emotion polarity into account. Moreover, it considers emotion mimicry during generation.
\paragraph{EmpDG}\footnote{\url{https://github.com/qtli/EmpDG}} \cite{li2019empgan}: This model detects nuanced emotion at word-level as a part of decoder inputs, and uses adversarial learning framework to involve user's feedback.

\paragraph{}In addition, four models that considers both the affection and cognition of empathy are selected:
\paragraph{KEMP}\footnote{\url{https://github.com/qtli/KEMP}} \cite{li-etal-2022-kemp} This model leverages external commonsense knowledge and emotional lexicon to understand and express emotion for empathetic response generation.
\paragraph{CEM}\footnote{\url{https://github.com/Sahandfer/CEM}} \cite{CEM2021} This model generates causes and effects of the user's latest mentioned experiences, and uses them to refine the context encoding for a better understanding of the user's situations and feelings.
\paragraph{RecEC$_{\text{soft}}$}\footnote{\url{https://github.com/A-Rain/EmpDialogue_RecEC}} \cite{gao-etal-2021-improving-empathetic}: This model pays more attention to emotion causes, detected from dialogue context, at word-level by a soft gated attention mechanism in the decoder.
\paragraph{GEE$_{\text{MIME}}$}\footnote{\url{https://github.com/skywalker023/focused-empathy}} \cite{Kim:2021:empathy}: This model uses a rational speech act framework to update the response generated by MIME to obtain the final response that focuses more on the emotion cause words in dialogue context.

\paragraph{}All above models, as well as ours, are built upon transformer backbone for a fair comparison.

\subsection{Implementation Details}
\paragraph{Our Model:} We implemented our model using PyTorch\footnote{\url{https://pytorch.org/}}, and trained it on a GPU of Nvidia GeForce RTX 3090. The token embeddings are initialized with 300-dimensional pre-trained Glove vectors \cite{pennington2014glove}, and shared between between the encoder, the CVGAE model, and the decoder. The hidden size $d$ is set as 300. The number of node number $|\mathcal{V}|$ is $800$, and the number of selected relationships $k$ is $512$ ($0.16\%$). 
Both the encoder layer number and the decoder layer number are $2$.
The batch size is set as $16$. When training the model, we use Adam optimizer \cite{kingma2014adam} and vary the learning rate following \citet{transformers}.

\paragraph{Comparison Models:} We implement GEE$_{\text{MIME}}$ under its official instructions, since only testing codes and instructions are provided by the authors. For the rest of the comparison models, we utilize their official codes released on GitHub.

\begin{table*}[t!]
    \centering
    \small
    \begin{tabular}{cl|c|cc|ccc}
    \specialrule{0.5pt}{0.5pt}{0.5pt}
    \multicolumn{2}{c|}{\textbf{Model}} & \textbf{PPL} & \textbf{BLEU-3} & \textbf{BLEU-4} & \textbf{P$_\text{BERT}$} &    \textbf{R$_\text{BERT}$} & \textbf{F$_\text{BERT}$}\\
    \specialrule{0.5pt}{0.5pt}{0.5pt}
    \multirow{3}{*}{\textbf{Affection}}
    & MoEL & 36.87$_{\pm0.97}$ & 4.53$_{\pm0.53}$ & 2.80$_{\pm0.32}$ & .499$_{\pm.008}$ & .467$_{\pm.007}$ & .480$_{\pm.006}$ \\
    & MIME & 37.88$_{\pm0.49}$ & 4.48$_{\pm0.15}$ & 2.71$_{\pm0.09}$ & .490$_{\pm.004}$ & .466$_{\pm.002}$ & .475$_{\pm.002}$ \\
    & EmpDG& 55.64$_{\pm3.78}$ & 3.64$_{\pm0.38}$ & 1.99$_{\pm0.22}$ & .475$_{\pm.007}$ & .458$_{\pm.008}$ & .465$_{\pm.004}$ \\
    \specialrule{0.5pt}{0.5pt}{0.5pt}
    \multirow{5}{*}{\makecell{\textbf{Affection}+\textbf{Cognition}}}
    & KEMP & 36.59$_{\pm0.45}$ & 4.13$_{\pm0.29}$ & 2.43$_{\pm0.15}$ & .484$_{\pm.005}$ & .460$_{\pm.004}$ & .470$_{\pm.005}$ \\
    & CEM  & 36.70$_{\pm0.44}$ & 3.55$_{\pm0.42}$  & 2.24$_{\pm0.24}$ & .498$_{\pm.001}$ & .461$_{\pm.006}$ & .477$_{\pm.004}$ \\
    & RecEC$_{\text{soft}}$& 149.3$_{\pm15.9}$ & 3.02$_{\pm0.15}$ & 1.62$_{\pm0.12}$ & .491$_{\pm.004}$ & .461$_{\pm.002}$ & .473$_{\pm.002}$ \\
    & GEE$_{\text{MIME}}$& - & 2.76$_{\pm0.18}$ & 1.50$_{\pm0.14}$ & .472$_{\pm.002}$ & .443$_{\pm.002}$ & .456$_{\pm.001}$ \\
    \cline{2-8}
    \specialrule{0pt}{0.5pt}{0.5pt}
    & CARE & \textbf{32.84}*$_{\pm0.23}$  & \textbf{4.88}*$_{\pm0.13}$ & \textbf{2.95}*$_{\pm0.06}$ & \textbf{.501}$_{\pm.004}$ & \textbf{.475}*$_{\pm.002}$ & \textbf{.486}*$_{\pm.003}$ \\
    \specialrule{0.5pt}{0.5pt}{0.5pt}
    \end{tabular}
    \caption{Automatic evaluation results in terms of \emph{PPL}, \emph{BLEU} and \emph{BERTScore}. For each method, we repeat five runs with different seeds. We display the average values of the results along with the standard deviations. The values marked with $*$ mean the results are statistically significant with $p<0.05$. The highest score in terms of each metric is in bold. The full automatic evaluation results can be found in \Cref{sec:appendix-auto}.}
    \label{tab:automatic}
\end{table*}

\subsection{Automatic Evaluation}\label{sec:automatic}
\paragraph{Metrics:} Three kinds of metrics are applied for automatic evaluation: (1) Perplexity (\textbf{PPL}), which measures the model's confidence in the response generation. (2) BLEU \cite{papineni2002bleu}, which estimates the matching between n-grams of the generated response and those of the golden response. We adopt \textbf{BLEU-3} and \textbf{BLEU-4}. (3) BERTScore \cite{zhang2019bertscore}, which computes the similarity for each token in the generated response with that in the golden response. We use its matching precision, recall and F1 score (\textbf{P$_\text{BERT}$}, \textbf{R$_\text{BERT}$}, and \textbf{F$_\text{BERT}$}). For perplexity, a lower score indicates a better performance; while, for the rest metrics, higher scores indicate better performances.

\paragraph{Annotation Statistics:} \Cref{tab:automatic} presents the automatic evaluation results, and the highest score in terms of each metric is in bold. For each model, we repeat five runs with different seeds, and compute the average values and standard deviations. In addition, values that are statistically significant with $p<0.05$ are marked with $*$. 

\paragraph{Results:} According to \Cref{tab:automatic}, our proposed model CARE outperforms the other models in terms of all metrics. 
The lowest perplexity score suggests that our proposed architecture is more confident in its generated responses than other models. The table does not present the perplexity score of GEE$_{\text{MIME}}$. This is because its generated token probability distribution depends on the mediate results of MIME and its emotion cause detector, and therefore PPL is less relevant to its core structure, i.e., rational speech act framework.
Highest BLEU and BERTScore scores indicate that our approach can generate more human-like responses by incorporating causality reasoning. 
Especially, all the above advantages are significant and stable, evident in high degrees of statistical significance and small standard deviations, respectively. 

\begin{table}[t!]
    \centering
    \small
    \begin{tabular}{cl|ccc}
    \specialrule{0.5pt}{0.5pt}{0.5pt}
    \multicolumn{2}{c|}{\textbf{Model}} & \textbf{Emp.} &    \textbf{Rel.} &    \textbf{Flu.} \\
    \specialrule{0.5pt}{0.5pt}{0.5pt}
    \multirow{3}{*}{\textbf{Affection}}
    & MoEL & 2.73 & 2.63 & 4.82 \\
    & MIME & 2.30 & 2.24 & 4.88  \\
    & EmpDG & 2.31 & 2.27 & 4.52 \\
     \specialrule{0.5pt}{0.5pt}{0.5pt}
    \multirow{5}{*}{\makecell{\textbf{Affection}\\+\textbf{Cognition}}}
    & KEMP & 2.26 & 2.18 & 4.81 \\
    & CEM & 2.77 & 2.70 & \textbf{4.93} \\
    & RecEC$_{\text{soft}}$ & 2.16 & 2.21 & 4.74 \\
    & GEE$_{\text{MIME}}$ & 1.75 & 1.75 & 4.78 \\
    \cline{2-5}
    \specialrule{0pt}{0.5pt}{0.5pt}
    & CARE & \textbf{2.83} & \textbf{2.79} & 4.86 \\
    \specialrule{0.5pt}{0.5pt}{0.5pt}
    \end{tabular}
    \caption{Results of human ratings in terms of \emph{Empathy}, \emph{Relevance} and \emph{Fluenct} on a $5-$point likert scale, where $5$ is the best. The highest scores are in bold. The fleiss's kappa is $0.41$ indicating a moderate level of agreement.}
    \label{tab:human}
\end{table}

\subsection{Human Ratings}\label{sec:human}
\paragraph{Metrics:} Although the automatic evaluation has provided useful information about models' performances, it cannot capture some features, such as empathy expression and contextual relevance. Therefore, following previous practices, we randomly sample $128$ conversations, and corresponding responses generated by different models for human ratings. We ask three human annotators to score each generated response from the following three aspects: (1) Empathy (\textbf{Emp.}), which measures whether the response understands user feelings and experiences. (2) Relevance (\textbf{Rel.}), which measures whether the response is on-topic and appropriate given the previous conversation. (3) Fluency (\textbf{Flu.}), which measures whether the response is fluent and its language is accurate. Each is on a $5-$point likert scale, where $5$ is the best. Then we compute the average value for each metric.

\paragraph{Annotation Statistics:} \Cref{tab:human} displays the human rating results, and the highest scores are in bold. We calculate Fleiss's kappa to measure inter-evaluator agreement of the human ratings. The result is $0.41$, indicating a moderate level of agreement among three annotators.

\paragraph{Results:} From these results, we can draw two conclusions. First, compared with most previous models, CARE achieves the highest scores in terms of \textbf{Emp.} and \textbf{Rel.}, and obtains relatively high \textbf{Flu.}. 
It indicates that our causality reasoning in an interdependent and simultaneous way indeed benefits empathetic expression and content relevance as we expect. 
Thanks to the reasoned causalities, CARE improves the understanding of user feelings and experiences.
In addition, the reasoning process enables the model to identify some reasonable user's feelings and experiences that are not explicitly mentioned in the previous conversation.
With such information, the model can show strong empathy in response, which is manifest in the case study. 
Second, models considering both affection and cognition (bottom half of the table) do not always outperform models merely considering affection (upper half of the table). This is also evident in \Cref{tab:ablation-automatic}, i.e., the automatic evaluation results. 
Although causality reasoning intuitively contributes to the understanding of user's feelings and experiences, inconsiderate reasoning can lead to one-sided understanding and low empathy.

\begin{table}[t!]
    \centering
    \small
    \begin{tabular}{l|c|c|c}
    \specialrule{0.5pt}{0.5pt}{0.5pt}
    {\textbf{Model Variant}} & \textbf{PPL} & \textbf{BLEU-3/4} & \textbf{P/R/F$_\text{BERT}$}\\
    \specialrule{0.5pt}{0.5pt}{0.5pt}
    w/o reasoning & 33.34 & 4.74/2.83 & .493/.473/.481  \\
    w/o condition & 33.23 & 4.74/2.83 & .501/.473/.485  \\
    \specialrule{0.5pt}{0.5pt}{0.5pt}
    Full model & 32.84 & 4.88/2.95 & .501/.475/.486 \\
    \specialrule{0.5pt}{0.5pt}{0.5pt}
    \end{tabular}
    \caption{Automatic evaluation results of the ablation study for CARE. The metrics are the same as those in \Cref{tab:automatic}. Similarly, we repeat five runs with different seeds, and display the average values. Its full automatic evaluation results can be found in \Cref{sec:appendix-auto}.}
    \label{tab:ablation-automatic}
\end{table}
\begin{table}[t!]
    \centering
    \small
    \begin{tabular}{l|ccc}
    \specialrule{0.5pt}{0.5pt}{0.5pt}
    {\textbf{Model Variant}} & \textbf{Emp.} &    \textbf{Rel.} &    \textbf{Flu.} \\
    \specialrule{0.5pt}{0.5pt}{0.5pt}
    w/o reasoning & 2.38 & 2.23 & 4.86 \\
    w/o condition & 2.60 & 2.47 & 4.87 \\
    \specialrule{0.5pt}{0.5pt}{0.5pt}
    Full model & 2.83 & 2.79 & 4.86 \\
    \specialrule{0.5pt}{0.5pt}{0.5pt}
    \end{tabular}
    \caption{Human rating results of the ablation study for CARE. The metrics are the same as those in \Cref{tab:human}.}
    \label{tab:ablation-human}
\end{table}

\subsection{Model Analysis}
In \cref{sec:automatic} and \cref{sec:human}, CARE has shown its superior performance. For deeper analyses of our model, we investigate its inner structures and functions.

\paragraph{Ablation Study} We propose two variant models to verify the contribution of reasoning and the reasoning condition in CARE:
\begin{itemize}
    \item \textbf{w/o reasoning}: We remove the CVGAE structure, and directly incorporate the prior causal graph into response generation. 
    \item \textbf{w/o condition}: We replace CVGAE with VGAE to eliminate the effect of the reasoning condition.
\end{itemize}
Results are shown in \Cref{tab:ablation-automatic} and \Cref{tab:ablation-human}, respectively. From \Cref{tab:ablation-automatic}, both variants achieve relatively high automatic evaluation metric scores. Moreover, the variant models surpass previous comparison models in \Cref{tab:automatic}.
It indicates that causalities can help models respond more like humans, given that both variants consider additional causalities between the user's experiences. However, both variants' performances in terms of human evaluation are relatively low. Accordingly, we can draw the following three conclusions:
\begin{itemize}
    \item Not all information in the golden response contributes to empathy. Although two variants have high automatic evaluation scores, they fail to achieve equally high human ratings. Such a phenomenon can also be clearly observed when comparing the performance of EmpDG and KMEP.
    \item Generated responses considering causalities not mentioned in dialogue history are more empathetic and relevant, which is supported theoretically by \citet{preston2002empathy}. It is reflected by higher \textbf{Emp.} and \textbf{Rel.} of \textbf{w/o condition} than \textbf{w/o reasoning}, given that \textbf{w/o condition} reasons plausible causalities not mentioned in dialogue history compared with \textbf{w/o reasoning}.
    \item Emotional and contextual information guides the model to reason causalities contributing to empathetic expression, given that the full model has higher Empathy and Relevant than \textbf{w/o condition}.
\end{itemize}

\begin{figure}
    \centering
    \includegraphics[width=\linewidth]{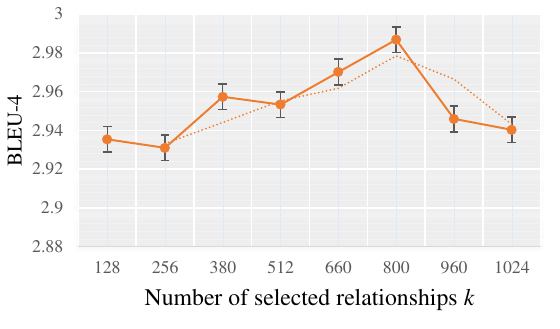}
    \caption{Model performs (\textbf{BLEU-4}) when we gradually increase the number of selected relationships $k$. The solid line and dot line represent BLEU-4 and two period moving average, respectively. For each $k$, we repeat five runs and compute the average BLEU-4.}
    \label{fig:relation}
\end{figure}
\paragraph{Analysis of \#SelectedRelationships $k$}
As shown in \Cref{fig:relation}, the performance of CARE with regard to BLEU-4 first rises and then drops as we increase the number of relationships infused in response generation $k$. It indicates that sufficient causalities benefit empathetic expression, but excess ones could involve noise and hurt empathy.

\begin{table}[t]
    \small
    \centering
        \begin{tabular}{m{.15\hsize}m{.75\hsize}}
        \specialrule{1pt}{.5pt}{.5pt}
        \multicolumn{2}{m{.95\hsize}}{
        \textbf{User Emotion Type}: apprehensive}\\
        \specialrule{0pt}{.5pt}{.5pt}
        \multicolumn{2}{m{.95\hsize}}{
        \textbf{User}: I had some hesitations when I was at the gym today, thought I wouldn't be able to handle the weights.} \\ 
        \specialrule{0pt}{.5pt}{.5pt}
        \multicolumn{2}{m{.95\hsize}}{
        \textbf{Bot}: How'd it turn out? Congrats on deciding to lead a healthier life, by the way.}\\
        \specialrule{0pt}{.5pt}{.5pt}
        \multicolumn{2}{m{.95\hsize}}{
        \textbf{User}: I decided to focus on the music that was blasting, just tried switching up what I play in the gym, it sounds like vikings are bringing me into battle} \\
        \specialrule{1pt}{.5pt}{.5pt}
        MoEL & That is great. I am sure you will do great!\\
        \specialrule{0pt}{.75pt}{.75pt}
        MIME & I am sure you will be fine. \\
        \specialrule{0pt}{.75pt}{.75pt}
        EmpDG & That is a great time to have! You should always feel the same way. \\
        \specialrule{0pt}{.75pt}{.75pt}
        KEMP & I am glad you are going through that.\\
        \specialrule{0pt}{.75pt}{.75pt}
        CEM & I am sure it will be fun!\\
        \specialrule{0pt}{.75pt}{.75pt}
        RecEC$_{\text{soft}}$ & That is a lot of money. How long did you do? \\
        \specialrule{0pt}{.75pt}{.75pt}
        GEE & I hate that. I hope you are okay.\\
        \specialrule{0pt}{.75pt}{.75pt}
        CARE & That is a \textit{good idea}. I hope you have a lot of \textit{confidence}.\\ 
        \specialrule{1pt}{.5pt}{.5pt}   
        \end{tabular}
    \caption{Responses generated by our method and previous empathetic response generation models. The content showing comprehension of feelings and experiences is highlighted in italic.}
    \label{tab:case}
\end{table}

\subsection{Case Study}
\Cref{tab:case} presents a case along with responses generated by our models and comparison models. From the table, CARE can respond more empathically to the user when compared with other models. Notably, CARE is able to show deep and considerate comprehension of the user's feelings and experiences in the response. For instance, it understands that the ``apprehensive'' emotion comes from lack of \textit{confidence} and the user has already proposed a quite effective solution (\textit{great idea}). 
\section{Conclusion}
In this paper, we propose to reason all plausible causalities in conversation interdependently and simultaneously for a deep understanding of the user's feelings and experiences in empathetic dialogue. Further, we turn the causality reasoning problem into a conditional graph generation task. Correspondingly, we design CVGAE, which uses latent variables for conditional structure prediction, and predicted future conversation content, to implement the reasoning. 
The reasoned causalities are infused into response generation for the final empathetic responses by a multi-source attention mechanism in the decoder.
This whole structure is named as CARE (\textbf{CA}usality \textbf{R}easoning for \textbf{E}mpathetic conversation).
Experimental results show that CARE outperforms prior methods in terms of both automatic and manual evaluations.

\section*{Limitations}
In this paper, we improve the model's empathy from the aspect of affection and cognition, especially the latter one. For this purpose, we incorporate reasoned causal knowledge into response generation. However, other knowledge, such as sentiment knowledge and commonsense knowledge, can also contribute to affection and cognition. KEMP \cite{li-etal-2022-kemp}, one of the comparison models in our experiment, has explored incorporating commonsense knowledge and sentiment knowledge into response generation. However, according to its model design, its use of knowledge is universal in chitchat conversations and is not aimed at empathetic expression. Therefore, it has low \textbf{Emp.} score as shown in \Cref{tab:human}. Therefore, it is worth exploring the connection between empathy and different types of knowledge. Besides, how to fuse different knowledge in one model for more empathetic responses is also a valuable problem.

\section*{Ethical Considerations}
The widely-used open-sourced \textsc{EmpatheticDialogues} \cite{empatheticdialogues} benchmark used in our experiment is collected through interaction with Amazon Mechanical Turk (MTurk). In this process, user privacy is protected, and no personal information is contained in the dataset. Therefore, we believe that our research work meets the ethics of EMNLP.

\section*{Acknowledgements}
This work was supported by the Research Grants Council of Hong Kong (PolyU/5204018, PolyU/15207920, PolyU/15207122) and National Natural Science Foundation of China (62076212).

\bibliography{custom}
\bibliographystyle{acl_natbib}

\appendix

\newpage
\section{Appendix}
\subsection{Automatic Evaluation}\label{sec:appendix-auto}
For our automatic evaluation, we modify codes\footnote{\url{https://github.com/ricsinaruto/dialog-eval}} for dialogue evaluations provided by \citet{Csaky:2019}. In addition, \Cref{tab:appendix} shows the full automatic evaluation results.
\begin{table*}[t!]
    \centering
    \small
    \begin{tabular}{l|c|cc|ccc}
    \specialrule{0.5pt}{0.5pt}{0.5pt}
    {\textbf{Model}} & \textbf{PPL} & \textbf{BLEU-3} & \textbf{BLEU-4} & \textbf{P$_\text{BERT}$} &    \textbf{R$_\text{BERT}$} & \textbf{F$_\text{BERT}$} \\
    \specialrule{0.5pt}{0.5pt}{0.5pt}
    
    MoEL & 36.42 & 4.74 & 2.89 & .489 & .469 & .477  \\
    MoEL & 37.43 & 5.23 & 3.23 & .503 & .476 & .487  \\
    MoEL & 37.94 & 4.60 & 2.82 & .511 & .470 & .487  \\
    MoEL & 37.12 & 3.61 & 2.25 & .490 & .456 & .470  \\
    MoEL & 35.45 & 4.48 & 2.83 & .486 & .469 & .475 \\
    \specialrule{0.5pt}{0.5pt}{0.5pt}
    MIME & 37.46 & 4.43 & 2.62 & .491 & .464 & .475  \\
    MIME & 38.36 & 4.21 & 2.60 & .485 & .464 & .472 \\
    MIME & 37.74 & 4.63 & 2.84 & .496 & .465 & .478  \\
    MIME & 37.40 & 4.60 & 2.79 & .490 & .469 & .477  \\
    MIME & 38.43 & 4.53 & 2.70 & .486 & .469 & .475  \\
    \specialrule{0.5pt}{0.5pt}{0.5pt}
    EmpDG& 59.01 & 3.42 & 1.76 & .477 & .465 & .469  \\
    EmpDG& 59.40 & 3.70 & 2.05 & .463 & .459 & .459 \\
    EmpDG& 56.41 & 3.03 & 1.72 & .483 & .444 & .461  \\
    EmpDG& 51.86 & 3.92 & 2.19 & .476 & .460 & .466  \\
    EmpDG& 51.54 & 4.13 & 2.25 & .478 & .464 & .469  \\
    \specialrule{0.5pt}{0.5pt}{0.5pt}
    KEMP & 36.87 & 4.44 & 2.61 & .477 & .465 & .469  \\
    KEMP & 37.24 & 3.86 & 2.27 & .463 & .459 & .459  \\
    KEMP & 36.26 & 3.70 & 2.22 & .483 & .444 & .461  \\
    KEMP & 36.17 & 4.28 & 2.50 & .476 & .460 & .466  \\
    KEMP & 36.42 & 4.36 & 2.54 & .478 & .464 & .469  \\
    \specialrule{0.5pt}{0.5pt}{0.5pt}
    CEM  & 36.47 & 3.99 & 2.47 & .498 & .468 & .480  \\
    CEM  & 37.12 & 2.93 & 1.87 & .497 & .451 & .470  \\
    CEM  & 36.61 & 4.03 & 2.54 & .500 & .468 & .482  \\
    CEM  & 36.13 & 3.40 & 2.18 & .500 & .460 & .477  \\
    CEM  & 37.15 & 3.39 & 2.15 & .497 & .460 & .475  \\
    \specialrule{0.5pt}{0.5pt}{0.5pt}
    RecEC$_{\text{soft}}$ & 139.78 & 3.25 & 1.80 & .493 & .463 & .475 \\
    RecEC$_{\text{soft}}$ & 136.33 & 3.08 & 1.70 & .494 & .461 & .475 \\
    RecEC$_{\text{soft}}$ & 173.85 & 3.03 & 1.62 & .484 & .462 & .470 \\
    RecEC$_{\text{soft}}$ & 157.11 & 2.88 & 1.55 & .492 & .458 & .472 \\
    RecEC$_{\text{soft}}$ & 139.60 & 2.85 & 1.46 & .491 & .462 & .475 \\
    \specialrule{0.5pt}{0.5pt}{0.5pt}
    GEE$_{\text{MIME}}$& - & 2.85 & 1.44 & .473 & .445 & .457 \\
    GEE$_{\text{MIME}}$& - & 2.83 & 1.61 & .475 & .443 & .457 \\
    GEE$_{\text{MIME}}$& - & 2.86 & 1.62 & .474 & .442 & .456 \\
    GEE$_{\text{MIME}}$& - & 2.40 & 1.26 & .470 & .441 & .453 \\
    GEE$_{\text{MIME}}$& - & 2.88 & 1.58 & .470 & .445 & .455 \\
    \specialrule{0.5pt}{0.5pt}{0.5pt}
    CAER & 32.64 & 5.03 & 2.99 & .507 & .477 & .490 \\
    CAER & 32.70 & 5.03 & 3.05 & .497 & .478 & .485 \\
    CAER & 33.16 & 4.87 & 2.95 & .503 & .476 & .487 \\
    CAER & 32.99 & 4.77 & 2.92 & .495 & .473 & .482 \\
    CAER & 32.69 & 4.70 & 2.86 & .503 & .473 & .486 \\
    \specialrule{0.5pt}{0.5pt}{0.5pt}
    w/o reasoning & 33.47 & 4.57 & 2.81 & .493 & .469 & .479 \\
    w/o reasoning & 33.39 & 4.80 & 2.82 & .484 & .472 & .476 \\
    w/o reasoning & 33.36 & 4.96 & 2.95 & .490 & .476 & .481 \\
    w/o reasoning & 33.18 & 4.70 & 2.79 & .500 & .474 & .485 \\
    w/o reasoning & 33.28 & 4.68 & 2.80 & .496 & .474 & .483 \\
    \specialrule{0.5pt}{0.5pt}{0.5pt}
    w/o condition & 32.56 & 4.70 & 2.85 & .497 & .475 & .484 \\
    w/o condition & 33.42 & 4.71 & 2.86 & .496 & .470 & .480 \\
    w/o condition & 32.86 & 4.97 & 2.96 & .508 & .475 & .489 \\
    w/o condition & 33.39 & 4.67 & 2.76 & .502 & .474 & .486 \\
    w/o condition & 33.94 & 4.65 & 2.75 & .503 & .470 & .484 \\
    \specialrule{0.5pt}{0.5pt}{0.5pt}
    \end{tabular}
    \caption{All automatic results from different methods with seed 0, 42, 1024, 1234 and 4096.}
    \label{tab:appendix}
\end{table*}

\subsection{Human Evaluation}
\begin{figure*}
    \centering
    \includegraphics[width=\textwidth]{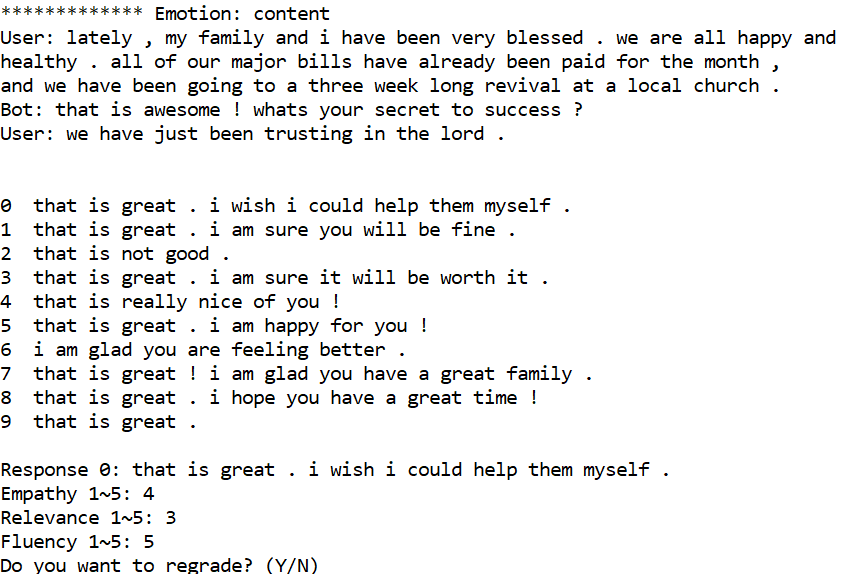}
    \caption{This is the user interface of the system for human ratings.}
    \label{fig:UI}
\end{figure*}

We implemented a system, as shown in \Cref{fig:UI}, for fair human ratings. For each case, we provide the previous dialogue turns, and the user emotion to the annotator. In addition, all responses generated by different models are displayed in a random order, thus the annotator cannot distinguish the source of each single response. 

Since human ratings are subjective, we provide some statements and classic examples as the reference for human evaluation.
\begin{itemize}
    \item \textit{Empathy}. We prefer responses with the following features: (1). Emotions, e.g., care, concern, and encourage. (2). Content, which shows interests in what the user cares. For instance, we prefer \textit{``Did you call the police?''} instead of \textit{``What movie?''} when the user says \textit{``It was stolen after the movie.''}.
    \item \textit{Relevance}. We prefer responses, based on which we can infer the topics in the previous dialogue content. 
    \item \textit{Fluency}. We reduce the marks if the following appears in a response: (1). Inappropriate (obvious features due to bad training) repetition, such as \textit{``I am sorry. I am sorry. I am sorry. I am''}. (2). Grammar mistakes, e.g., misuse of personal pronouns and tense. (3). Conflicting contents, such as \textit{``I can understand you. I cannot understand you.''}. 
\end{itemize}
Moreover, we encourage annotators to compare different responses in mind before grading each response.

\end{document}